\documentclass{article}
\pdfoutput=1

\usepackage[preprint, nonatbib]{neurips_data_2023}

\usepackage[utf8]{inputenc} 
\usepackage[T1]{fontenc}    
\usepackage{url}            
\usepackage{booktabs}       
\usepackage{amsfonts}       
\usepackage{nicefrac}       
\usepackage{microtype}      
\usepackage{xcolor}         
\usepackage{tabu}         
\usepackage{xspace,mfirstuc,tabulary}

\usepackage{multirow}
\usepackage{array, caption, floatrow, makecell, booktabs}
\usepackage{wrapfig}
\usepackage{floatrow}
\usepackage{comment}
\usepackage{footnote}
\usepackage{enumitem}
\usepackage{amsmath}
\usepackage{accents}

\usepackage{graphicx}
\usepackage{bbding}
\usepackage{ntheorem}
\usepackage{enumitem}
\usepackage{oplotsymbl}
\usepackage{multirow}
\usepackage{bbm}
\usepackage{adjustbox}
\usepackage{subcaption}
\usepackage{caption}
\usepackage{multicol}
\usepackage{wrapfig}

\usepackage{tabu}

\usepackage{multirow,array}
\usepackage{color, colortbl}
\definecolor{Gray}{gray}{0.93}

\usepackage{booktabs}
\usepackage{pifont}

\usepackage[pagebackref=true,breaklinks=true,colorlinks,bookmarks=false]{hyperref}

\usepackage{floatrow}
\newlength\savewidth
\newcommand\paperurl[1]{{\footnotesize{\color{blue}{\url{#1}}}}}

\makeatletter
\DeclareRobustCommand\onedot{\futurelet\@let@token\@onedot}
\def\@onedot{\ifx\@let@token.\else.\null\fi\xspace}

\def\ie{\emph{i.e}\onedot}

\makeatother

\definecolor{rose}{HTML}{003472}

\newcommand{\sota}[1]{\textcolor{black}{$\textbf{#1}$}}

\title{DINO-X: A Unified Vision Model for \\Open-World Object Detection and Understanding}

\author{%
 \textbf{IDEA Research Team} \\
  \\
  International Digital Economy Academy (IDEA), IDEA Research \\
  \url{https://deepdataspace.com/home}
}

\begin{document}

\maketitle

\vspace{-1.0cm}

\begin{figure}[ht!]
\centering
\includegraphics[width=0.98\textwidth,keepaspectratio]{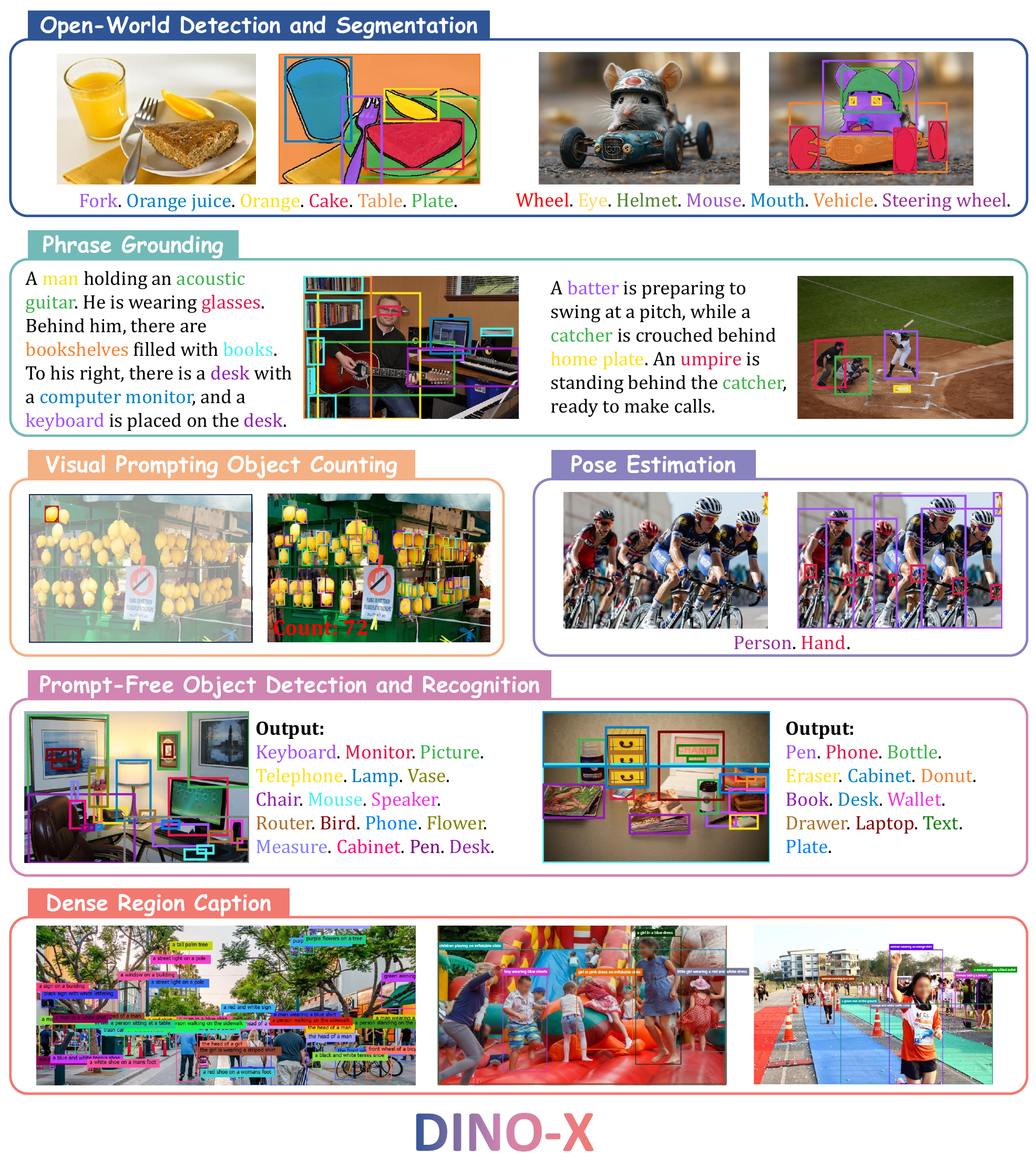}
\vspace{-0.2cm}
 \caption{DINO-X is a unified object-centric vision model which supports various open-world perception and object-level understanding tasks, including Open-World Object Detection and Segmentation, Phrase Grounding, Visual Prompt Counting, Pose Estimation, Prompt-Free Object Detection and Recognition, Dense Region Caption, etc.}
 \label{fig:dinox_teaser}
\end{figure}

\begin{abstract}

In this paper, we introduce {DINO-X}, which is a unified object-centric vision model developed by IDEA Research with the best open-world object detection performance to date. DINO-X employs the same Transformer-based encoder-decoder architecture as Grounding DINO 1.5~\cite{GroundingDINO1.5} to pursue an object-level representation for open-world object understanding. To make long-tailed object detection easy, DINO-X extends its input options to support text prompt, visual prompt, and customized prompt. With such flexible prompt options, we develop a universal object prompt to support \textit{prompt-free} open-world detection, making it possible to detect anything in an image without requiring users to provide any prompt.
To enhance the model's core grounding capability, we have constructed a large-scale dataset with over 100 million high-quality grounding samples, referred to as {Grounding-100M}, for advancing the model's open-vocabulary detection performance. Pre-training on such a large-scale grounding dataset leads to a foundational object-level representation, which enables DINO-X to integrate multiple perception heads to simultaneously support multiple object perception and understanding tasks, including detection, segmentation, pose estimation, object captioning, object-based QA, etc. DINO-X encompasses two models: the Pro model, which provides enhanced perception capabilities for various scenarios, and the Edge model, which is optimized for faster inference speed and better suited for deployment on edge devices. Experimental results demonstrate the superior performance of DINO-X. Specifically, the DINO-X Pro model achieves $56.0$ AP, $59.8$ AP, and $52.4$ AP on the COCO, LVIS-minival, and LVIS-val zero-shot object detection benchmarks, respectively. Notably, it scores $63.3$ AP and $56.5$ AP on the rare classes of LVIS-minival and LVIS-val benchmarks, improving the previous SOTA performance by $5.8$ AP and $5.0$ AP. Such a result underscores its significantly improved capacity for recognizing long-tailed objects. Our demo and API will be released at \href{https://github.com/IDEA-Research/DINO-X-API}{https://github.com/IDEA-Research/DINO-X-API}.

\end{abstract}

\section{Introduction}

In recent years, object detection has gradually evolved from closed-set detection models~\cite{DINO, MaskDINO, Mask2Former} to open-set detection models~\cite{GroundingDINO, GLIP, GLIPv2}, which can identify objects corresponding to user-provided prompt. Such models have numerous practical applications, such as enhancing the adaptability of robots in dynamic environments, assisting autonomous vehicles in rapidly locating and reacting to new objects, improving the perceptual capabilities of multimodal large language models (MLLMs), reducing their hallucinations, and increasing the reliability of their responses. 

In this paper, we introduce DINO-X, which is a unified object-centric vision model developed by IDEA Research with the best open-world object detection performance to date. Building upon Grounding DINO 1.5~\cite{GroundingDINO1.5}, DINO-X employs the same Transformer encoder-decoder architecture and adopts open-set detection as its core training task.  

To make long-tailed object detection easy, DINO-X incorporates a more comprehensive prompt design at the model's input stage. Traditional text prompt-only models~\cite{GroundingDINO, GroundingDINO1.5, GLIP}, while having made great progress, still struggle to cover a sufficient range of long-tailed detection scenarios due to the difficulty of collecting sufficiently diverse training data to cover various applications. To overcome this shortage, in DINO-X, we extend the model architecture to support the following three types of prompts. (1) {Text Prompt}: This involves identifying desired objects based on user-provided text input, which can cover most of the detection scenarios. (2) {Visual Prompt}: Beyond text prompts, DINO-X also supports visual prompts as in T-Rex2~\cite{TRex2}, further covering detection scenarios that cannot be well described by text alone. (3) {Customized Prompt}: To enable more long-tailed detection problems, we particularly introduce customized prompt in DINO-X, which can be implemented as either pre-defined or user-tuned prompt embeddings for customized needs. Through prompt-tuning, we can create domain-customized prompts for different domains or function-specific prompts to address various functional needs. For instance, in DINO-X, we develop a universal object prompt to support \textit{prompt-free} open-world object detection, making it possible to detect any objects in a given image without requiring users to provide any prompt. 

To achieve a strong grounding performance, we collected and curated over 100 million high-quality grounding samples from diverse sources, termed as {Grounding-100M}. 
Pre-training on such a large-scale grounding dataset leads to a foundational object-level presentation, which enables DINO-X to integrate multiple perception heads to simultaneously support multiple object perception and understanding tasks. 
Beyond the box head for object detection, DINO-X has implemented three additional heads: (1) {Mask Head} for predicting segmentation masks for the detected objects, (2) {Keypoint Head} for predicting more semantically meaningful keypoint for specific categories, and (3) {Language Head} for generating fine-grained descriptive captions for each detected object. By integrating these heads, DINO-X could provide more detailed object-level understanding of an input image. In Figure~\ref{fig:dinox_teaser}, we list various examples to illustrate the object-level vision tasks supported by DINO-X.

\begin{figure}[ht!]
\centering
 \includegraphics[width=1.0\textwidth,keepaspectratio]{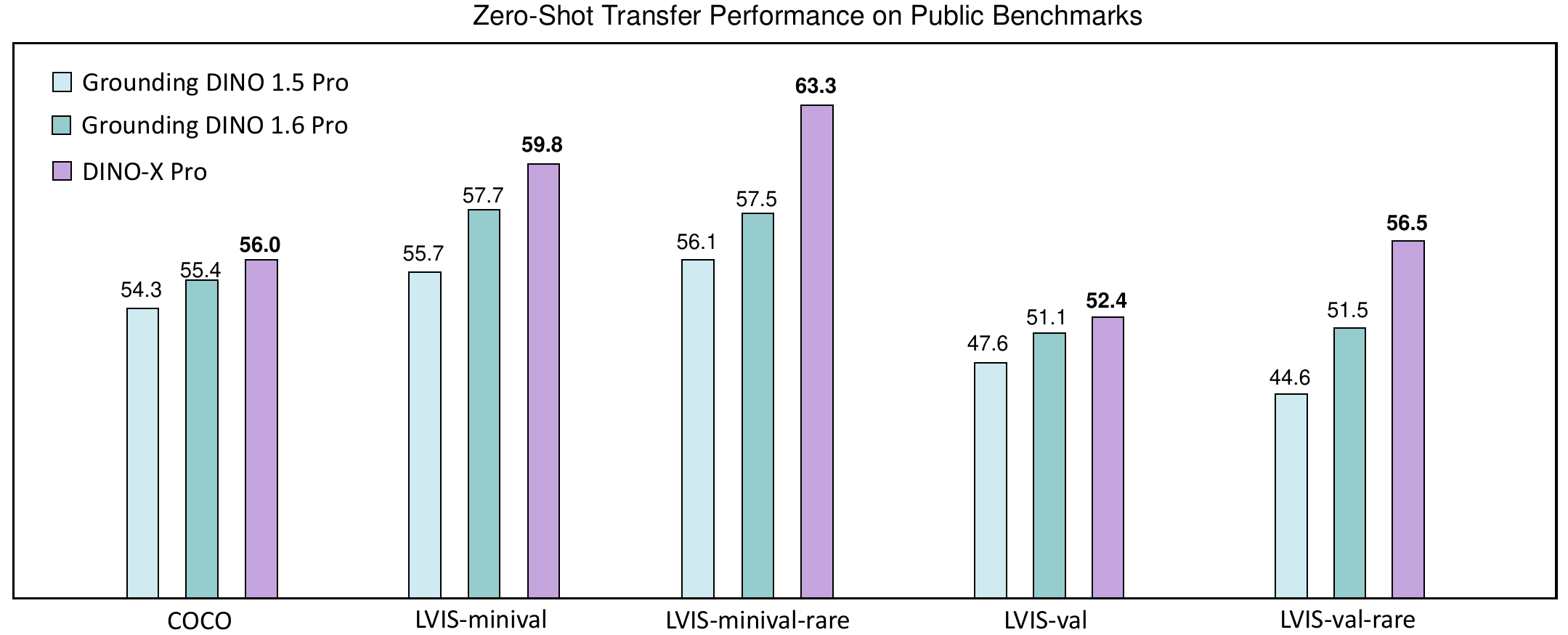}
 \caption{{DINO-X Pro} zero-shot performance on public detection benchmarks. Comparing with Grounding DINO 1.5 Pro and Grounding DINO 1.6 Pro, DINO-X Pro achieves new state-of-the-art (SOTA) performance on COCO, LVIS-minival, and LVIS-val zero-shot benchmarks. Furthermore, it outperforms other models with larger margins in detecting rare classes of objects on LVIS-minival and LVIS-val, demonstrating its exceptional capability of recognizing long-tailed objects.}
 \label{fig:dino_x_performance}
 
\end{figure}

Similar to Grounding DINO 1.5, DINO-X also encompasses two models: the DINO-X Pro model, which provides enhanced perception capabilities for various scenarios, and the DINO-X Edge model, which is optimized for faster inference speed and better suited for deployment on edge devices. Experimental results demonstrate the superior performance of DINO-X. As illustrated in Figure~\ref{fig:dino_x_performance}, our DINO-X Pro model achieves $56.0$ AP, $59.8$ AP, and $52.4$ AP on the COCO, LVIS-minival, and LVIS-val zero-shot transfer benchmarks, respectively. Notably, it scores $63.3$ AP and $56.5$ AP on the rare classes of the LVIS-minival and LVIS-val benchmarks, showing improvements of $5.8$ AP and $5.0$ AP over Grounding DINO 1.6 Pro, and $7.2$ AP and $11.9$ AP over Grounding DINO 1.5 Pro, highlighting its significantly improved ability to recognize long-tailed objects.
  
\section{Approach}

\begin{figure}[ht!]
\centering
 \includegraphics[width=1.0\textwidth,keepaspectratio]{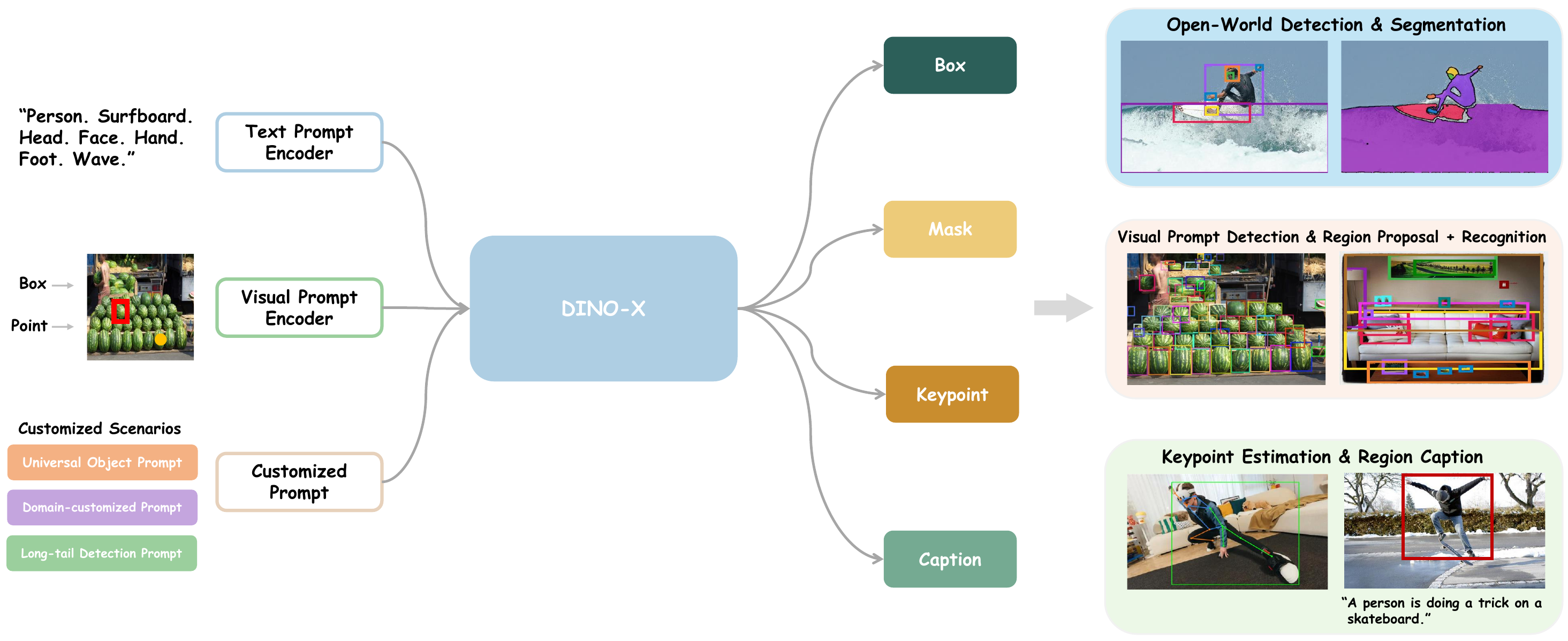}
 \caption{DINO-X is designed to accept text prompt, visual prompt, and customized prompt, and is capable of simultaneously generating outputs ranging from coarse-level representations, such as bounding boxes, to fine-grained details, including masks, keypoints, and object captions.}
 \label{fig:dino_x_framework}
\end{figure}

\subsection{Model Architecture}

The overall framework of DINO-X is shown in Fig.~\ref{fig:dino_x_framework}. Following Grounding DINO 1.5, we also develop two variants of DINO-X models: a more powerful and comprehensive "Pro" version, {DINO-X Pro}, as well as a faster "Edge" version, termed {DINO-X Edge}, which will be introduced in details in Sections \ref{dino_x_pro} and \ref{dino_x_edge}, respectively.

\subsubsection{DINO-X Pro}
\label{dino_x_pro}

The core architecture of the DINO-X Pro model is similar to Grounding DINO 1.5~\cite{GroundingDINO1.5}. We utilize a pre-trained ViT~\cite{EVA-02} model as its primary vision backbone and employ a deep early fusion strategy during the feature extraction stage. Different from Grounding DINO 1.5, to further extend the model's capability of detecting long-tailed objects, we have broadened the prompt support in DINO-X Pro at the input stage. Besides text prompts, we extend DINO-X Pro to also support visual prompts and customized prompts to cover various detection needs. Text prompts can cover the majority of object detection scenarios commonly encountered in daily life, while visual prompts enhance the model's detection capability in situations where text prompts fall short due to data scarcity and descriptive limitations~\cite{TRex2}. Customized prompts are defined as a series of specialized prompts that can be fine-tuned through prompt-tuning~\cite{PromptTuning} techniques to expand the model's ability to detect objects in more long-tailed, domain-specific, or function-specific scenarios without compromising other capabilities. 
By performing large-scale grounding pre-training, we obtain a foundational object-level representation from the encoder output of DINO-X. Such a robust representation enables us to seamlessly support multiple object perception or understanding tasks by introducing different perception heads. As a result, DINO-X is capable of generating outputs across different semantic levels, ranging from coarse-level, such as bounding boxes, to more fine-grained level, including masks, keypoints, and object captions. 

We will first introduce the supported prompts in DINO-X in the following paragraphs.

\paragraph{Text Prompt Encoder:} 
Both Grounding DINO~\cite{GroundingDINO} and Grounding DINO 1.5~\cite{GroundingDINO1.5} employ BERT~\cite{BERT} as text encoder. However, the BERT model is trained solely on text data, which limits its effectiveness for  perception tasks requiring multimodal alignment, such as open-world detection. Therefore, in DINO-X Pro, we utilize a pre-trained CLIP~\cite{MetaCLIP} model as our text encoder, which has pre-trained on extensive multimodal data, thereby further enhancing the model's training efficiency and performance across various open-world benchmarks.

\paragraph{Visual Prompt Encoder:}
We adopt the visual prompt encoder from T-Rex2~\cite{TRex2}, integrating it to enhance object detection by utilizing user-defined visual prompts in both box and point formats. These prompts are converted into position embeddings using a sine-cosine layer and then projected into a unified feature space. The model separates box and point prompts using different linear projections. Then we employ the same multi-scale deformable cross-attention layers as in T-Rex2 to extract visual prompt features from multi-scale feature maps, conditioned on the user-provided visual prompts.

\paragraph{Customized Prompt:} In practical use cases, it is common to encounter the need for fine-tuning models for customized scenarios. In DINO-X Pro, we define a series of specialized prompts, termed customized prompt, which can be fine-tuned through prompt-tuning~\cite{PromptTuning} techniques to cover more long-tailed, domain-specific, or function-specific scenarios in a {resource-efficient} and {cost-effective} manner without compromising other capabilities. For instance, we developed a {universal object prompt} to support \textit{prompt-free} open-world detection, making it possible to detect any objects within an image, thereby expanding its potential applications in areas such as screen parsing~\cite{OmniParser}, etc.

Given an input image and a user-provided prompt, no matter it is textual, visual, or a customized prompt embedding, DINO-X performs deep feature fusion between the prompt and the visual features extracted from the input image and then apply different heads for different perception tasks. More specifically, the implemented heads are introduced in the following paragraphs.

\paragraph{Box Head:} Following Grounding DINO~\cite{GroundingDINO}, we adopt the language-guided query selection module to select features that are most relevant to the input prompt as decoder object queries. Each query is then fed into the Transformer decoder and updated layer-by-layer, followed by a simple MLP layer that predicts the corresponding bounding box coordinates for each object query. Similar to Grounding DINO, we employ L1 loss and G-IoU~\cite{G-IoU} loss for bounding box regression, while using contrastive loss to align each object query with the input prompt for classification.

\paragraph{Mask Head:} Following the core design of Mask2Former~\cite{Mask2Former} and Mask DINO~\cite{MaskDINO}, we construct the pixel embedding map by fusing the 1/4 resolution backbone feature and the upsampled 1/8 resolution feature from the Transformer encoder. Then we perform dot-product between each object query from the Transformer decoder and the pixel embedding map to get the mask output of the query. In order to improve the training efficiency, the 1/4 resolution feature map from the backbone was only used in mask prediction. And we also follow~\cite{PointRend, Mask2Former} to only compute the mask loss for sampled points in the final mask loss calculation.

\paragraph{Keypoint Head:} The keypoint head takes keypoint-related detection outputs from DINO-X, e.g. person or hand, as input and utilize a separate decoder to decode object keypoints. Each detection output is treated as a query and expanded into a number of keypoints, which are then sent to multiple deformable Transformer decoder layers to predict the desired keypoint positions and their visibilities. This process can be regarded as a simplified ED-Pose~\cite{yang2023explicit} algorithm, which does not need to consider the object detection task but only focuses on keypoint detection. In DINO-X, we instantiate two keypoint heads for person and hand, which have 17 and 21 pre-defined keypoints, respectively.

\begin{figure}[ht!]
\centering
 \includegraphics[width=0.8\textwidth,keepaspectratio]{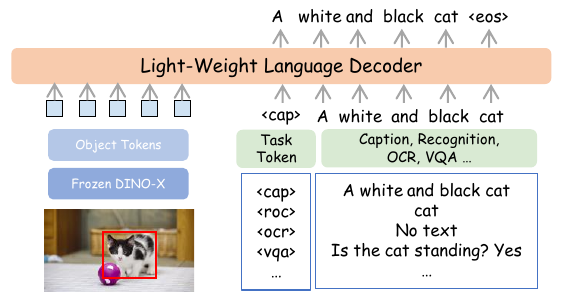}
 \caption{The detailed design of language head in DINO-X. It involves using a frozen DINO-X to extract object tokens, and a linear projection aligns its dimensions with the text embeddings. The lightweight language decoder then integrates these object and task tokens to generate response outputs in an autoregressive manner. The task tokens equip the language decoder with the capability of tackling different tasks.}
 \label{fig:dinox_cap}
\end{figure}

\paragraph{Language Head:}
The language head is a task-promptable generative small language model to enhance DINO-X's ability to comprehend regional context and perform perception tasks beyond localization, such as object recognition, region captioning, text recognition, and region-based visual question answering (VQA). The architecture of our model is depicted in Figure \ref{fig:dinox_cap}. For any detected object from DINO-X, we first extract its region features from the DINO-X backbone features using the RoIAlign \cite{he2017mask} operator, combined with its query embedding to form our object tokens. Then, we apply a simple linear projection to ensure their dimensions aligned with the text embedding. The lightweight language decoder integrates these regional representations with task tokens to generate outputs in an auto-regressive manner. The learnable task tokens empower the language decoder to handle a variety of tasks.

\subsubsection{DINO-X Edge}
\label{dino_x_edge}

Following Grounding DINO 1.5 Edge~\cite{GroundingDINO1.5}, DINO-X Edge also utilizes EfficientViT~\cite{cai2023efficientvit} as backbone for efficient feature extraction and incorporates a similar Transformer encoder-decoder architecture. To further enhance DINO-X Edge model's performance and computational efficiency, we employ several improvements to the model architecture and training techniques in the following aspects:

\paragraph{Stronger Text Prompt Encoder:} To achieve more effective region-level multi-modal alignment, DINO-X Edge adopts the same CLIP text encoder as our Pro model. In practice, text prompt embeddings can be pre-computed for most cases and do not affect the inference speed of the visual encoder and decoder. Using a stronger text prompt encoder generally leads to better results.

\paragraph{Knowledge Distillation:} 
In DINO-X Edge, we distill the knowledge from the Pro model to enhance the Edge model's performance. Specifically, we utilize both feature-based distillation and response-based distillation, which align the feature and prediction logits between the Edge model and the Pro model, respectively.
This knowledge transfer enables DINO-X Edge to achieve a stronger zero-shot capability compared to Grounding DINO 1.6 Edge.

\paragraph{Improved FP16 Inference:}
We employ a normalization technique for floating-point multiplication, enabling model quantization into FP16 without compromising accuracy.
This results in an inference speed of $20.1$ FPS, a $33\%$ increase from $15.1$ FPS compared to Grounding DINO 1.6 Edge, and a $87\%$ improvement from $10.7$ FPS compared to Grounding DINO 1.5 Edge.

\section{Dataset Construction and Model Training}

\paragraph{Data Collection:} To ensure the core open-vocabulary object detection capability, we developed a high-quality and semantic-rich grounding dataset, which consists of over 100 million images collected from the web, termed {Grounding-100M}. We used the training data from T-Rex2 with some additional industrial scenario data for visual prompt-based grounding pre-training.
We used open-source segmentation models, such as SAM~\cite{SAM} and SAM2~\cite{SAM2}, to generate pseudo mask annotations for a portion of the Grounding-100M dataset, which serves as the main training data for our mask head. we sampled a subset of high-quality data from the Grounding-100M dataset and utilized their box annotations as our \textit{prompt-free} detection training data. We also collected over 10 million region understanding data, covering object recognition, region captioning, OCR, and region-level QA scenarios for language head training.

\paragraph{Model Training:} To overcome the challenge of training multiple vision tasks, we adopt a two-stage strategy. In the first stage, we conducted joint training for text-prompt-based detection, visual-prompt-based detection, and object segmentation. In this training phase, we did not incorporate any images or annotations from COCO~\cite{COCO}, LVIS~\cite{LVIS}, and V3Det~\cite{V3Det} datasets, so that we can evaluate the model's zero-shot detection performance on these benchmarks. Such a large-scale grounding pre-training ensures an outstanding open-vocabulary grounding performance of DINO-X and results in a foundational object-level representation. In the second stage, we froze the DINO-X backbone and added two keypoint heads (for person and hand) and a language head, each being trained separately. By adding more heads, we greatly expand DINO-X's ability to perform more fine-grained perception and understanding tasks, such as pose estimation, region captioning, object-based QA, etc. Subsequently, we leveraged prompt-tuning techniques and trained a universal object prompt, allowing for prompt-free any-object detection while preserving the model's other capabilities. Such a two-stage training approach has several advantages: (1) it ensures that the model's core grounding capability is not affected by introducing new abilities, and (2) it also validates that large-scale grounding pre-training can serve as a robust foundation for an object-centric model, allowing for seamless transfer to other open-world understanding tasks.

\section{Evaluation}

In this section, we compare the various capabilities of our DINO-X series model with its related works. The best and the second best results are indicated in \textbf{bold} and with \underline{underline} 

\subsection{DINO-X Pro}

\subsubsection{Open-World Detection and Segmentation}

\paragraph{Evaluation on Zero-Shot Object Detection and Segmentation Benchmarks:} Following Grounding DINO 1.5 Pro~\cite{GroundingDINO1.5}, we evaluate the zero-shot object detection and segmentation capability of DINO-X Pro on the COCO~\cite{COCO} benchmark, which includes 80 common categories, and the LVIS benchmark, which features a richer and more extensive long-tail distribution of categories. As shown in Table~\ref{tab:dino-x-pro-zero-shot}, DINO-X Pro shows a significant performance improvement compared to previous state-of-the-art methods. Specifically, on the COCO benchmark, DINO-X Pro achieves an increase of $1.7$ box AP and $0.6$ box AP compared to Grounding DINO 1.5 Pro and Grounding DINO 1.6 Pro, respectively. On the LVIS-minival and LVIS-val benchmarks, DINO-X Pro achieves $59.8$ box AP and $52.4$ box AP, respectively, surpassing the previously best-performing Grounding DINO 1.6 Pro model by $2.0$ AP and $1.1$ AP, respectively. Notably, for the detection performance on LVIS rare classes, DINO-X achieves $63.3$ AP on LVIS-minival and $56.5$ AP on LVIS-val, significantly surpassing the previous SOTA Grounding DINO 1.6 Pro model by $5.8$ AP and $5.0$ AP, respectively, demonstrating the exceptional capability of DINO-X in long-tailed object detection scenarios. In terms of segmentation metrics, we compared DINO-X with the most commonly used general segmentation model, Grounded SAM~\cite{GroundedSAM} series, on the COCO and LVIS zero-shot instance segmentation benchmarks. Using Grounding DINO 1.5 Pro for zero-shot detection and SAM-Huge~\cite{SAM} for segmentation, Grounded SAM achieves the best zero-shot performance on the LVIS instance segmentation benchmarks. DINO-X achieves mask AP scores of $37.9$, $43.8$, and $38.5$ on the COCO, LVIS-minival, and LVIS-val zero-shot instance segmentation benchmarks, respectively. Compared to Grounded SAM, there is still a notable performance gap for DINO-X to catch up, which shows the challenge of training a unified model for multiple tasks. Nevertheless, DINO-X significantly improves the segmentation efficiency by generating corresponding masks for each region without requiring multiple complex inference steps. We will further optimize the performance of the mask head in our future work.

\begin{table}[!ht]
\setlength{\tabcolsep}{3pt}
\renewcommand{\arraystretch}{1.25}
\centering
\scriptsize
\resizebox{1.00\linewidth}{!}{
\begin{tabular}{lccccccccccccccccccc}
\toprule
\multirow{3}*{Method} & \multirow{3}*{Backbone} & \multicolumn{2}{c}{COCO-val} & \multicolumn{8}{c}{LVIS-minival} & \multicolumn{8}{c}{LVIS-val} \\
\cmidrule(lr){5-12} \cmidrule(lr){13-20}
 &  &   & &  \multicolumn{4}{c}{Box AP} & \multicolumn{4}{c}{Mask AP} & \multicolumn{4}{c}{Box AP} & \multicolumn{4}{c}{Mask AP} \\
 \cmidrule(lr){5-8} \cmidrule(lr){9-12} \cmidrule(lr){13-16} \cmidrule(lr){17-20} 
& & $\text{AP}_{box}$ & $\text{AP}_{mask}$ & $\text{AP}_{all}$ & $\text{AP}_{r}$ & $\text{AP}_{c}$ & $\text{AP}_{f}$ & $\text{AP}_{all}$ & $\text{AP}_{r}$ & $\text{AP}_{c}$ & $\text{AP}_{f}$ & $\text{AP}_{all}$ & $\text{AP}_{r}$ & $\text{AP}_{c}$ & $\text{AP}_{f}$ & $\text{AP}_{all}$ & $\text{AP}_{r}$ & $\text{AP}_{c}$ & $\text{AP}_{f}$ \\
\midrule
\multicolumn{20}{c}{\textit{Supervised Model (Pretraining data includes COCO, LVIS, etc.)}}\\
\midrule
GLIPv2~\cite{GLIPv2} & Swin-H & \textcolor{gray}{60.6} & - & \textcolor{gray}{50.1} & - & - & - & - & - & - & - & - & - & - & - & - & - & - & - \\
Grounding DINO~\cite{GroundingDINO} & Swin-L & \textcolor{gray}{60.7} & - & \textcolor{gray}{33.9} & \textcolor{gray}{22.2} & \textcolor{gray}{30.7} & \textcolor{gray}{38.8} & - & - & - & - & - & - & - & - & - & - & - & - \\
APE (B)~\cite{APE} & ViT-L & \textcolor{gray}{57.7} & \textcolor{gray}{48.6} & \textcolor{gray}{62.5} & - & - & - & \textcolor{gray}{55.4} & - & - & - & \textcolor{gray}{57.0} & - & - & - & \textcolor{gray}{50.5} & - & - & - \\
APE (D)~\cite{APE} & ViT-L & \textcolor{gray}{58.3} & \textcolor{gray}{49.3} & \textcolor{gray}{64.7} & - & - & - & \textcolor{gray}{57.5} & - & - & - & \textcolor{gray}{59.6} & - & - & - & \textcolor{gray}{53.0} & - & - & - \\
GLEE-Pro~\cite{GLEE} & ViT-L & \textcolor{gray}{62.0} & \textcolor{gray}{54.2} & - & - & - & - & - & - & - & - & \textcolor{gray}{55.7} & \textcolor{gray}{49.2} & - & - & \textcolor{gray}{49.9} & \textcolor{gray}{44.3} & - & - \\
DINOv~\cite{li2023visual} & Swin-T & \textcolor{gray}{47.0} & \textcolor{gray}{42.7} & - & - & - & - & - & - & - & - & - & - & - & - & - & - & - & -  \\
DINOv~\cite{li2023visual} & Swin-L & \textcolor{gray}{54.2} & \textcolor{gray}{50.4} & - & - & - & - & - & - & - & - & - & - & - & - & - & - & - & - \\
\midrule
\multicolumn{20}{c}{\textit{Zero-shot Transfer Model}} \\
\midrule
OWL-ViT~\cite{minderer2022simple}  & ViT-L  & 42.2 & - & - & - & - & - & - & - & - & - & 34.6 & 31.2 & - & - & - & - & - & - \\
MDETR~\cite{kamath2021mdetr}  & RestNet101  & -  & - & 22.5 & 7.4 & 22.7 & 25.0 & - & - & - & - & - & - & - & - & - & - & - & - \\
GLIP~\cite{GLIP}  & Swin-L  & 49.8 & -  & 37.3 & 28.2 & 34.3 & 41.5 & - & - & - & - & 26.9 & 17.1 & 23.3 & 35.4 & - & - & - & -  \\
Grounding DINO~\cite{GroundingDINO} & Swin-T & 48.4 & - & 27.4 & 18.1 & 23.3 & 32.7 & - & - & - & - & - & - & - & - & - & - & - & - \\
Grounding DINO~\cite{GroundingDINO} & Swin-L  & 52.5 & - & - & - & - & - & - & - & - & - & - & - & - & - & - & - & - & - \\
OpenSeeD~\cite{zhang2023simple} & Swin-L  & - & - & 23.0 & - & - & - & 21.0 & - & - & -  & - & - & - & - & - & - & - & - \\
UniDetector~\cite{wang2023detecting} & ResNet50  & - & - & - & - & - & - & - & - & - & - & 19.8 & 18.0 & 19.2 & 21.2 & - & - & - & - \\
OmDet-Turbo-B~\cite{OmDet_Turbo} & ConvNeXt-B   & 53.4 & - & 34.7 & - & - & - & - & - & -  & - & - & - & - & - & - & - & - & - \\
OWL-ST~\cite{OWLViTV2}  & CLIP L/14 & - & - & 40.9 & 41.5 & - & - & - & - & - & - & 35.2 & 36.2 & - & - & - & - & - & - \\
MQ-GLIP~\cite{xu2024multi}  & Swin-L & - & - & 43.4 & 34.5 & 41.2 & 46.9 & - & - & - & - & 34.7 & 26.9 & 32.0 & 41.3 & - & - & - & -\\
MM-Grounding-DINO~\cite{MM-Grounding-DINO}  & Swin-T  & 50.4 & - & 41.4 & 34.2 & 37.4 & 46.2 & - & - & - & - & 31.9 & 23.6 & 27.6 & 40.5 & - & - & - & -\\
MM-Grounding-DINO~\cite{MM-Grounding-DINO} & Swin-L  & 53.0 & - & - & - & - & - & - & - & -  & - & - & - & - & - & - & - & - & - \\
DetCLIP~\cite{DetCLIP}  & Swin-L  & - & -  & 38.6 & 36.0 & 38.3 & 39.3 & - & - & - & - & 28.4 & 25.0 & 27.0 & 31.6 & - & - & - & - \\
DetCLIPv2~\cite{DetCLIPv2}  & Swin-L  & - & - & 44.7 & 43.1 & 46.3 & 43.7 & - & - & - & - & 36.6 & 33.3 & 36.2 & 38.5 & - & - & - & - \\
DetCLIPv3~\cite{DetCLIPv3} & Swin-L & - & - & 48.8 & 49.9 & 49.7 & 47.8 & - & - & - & - & 41.4 & 41.4 & 40.5 & 42.3 & - & - & - & - \\
YOLO-World~\cite{YOLO-World} & YOLOv8-L  & 45.1 & - & 35.4 & 27.6 & 34.1 & 38.0  & - & - & - & - & - & - & - & - & - & - & - & -  \\
OV-DINO~\cite{OV-DINO} & Swin-T  & 50.2 & - & 40.1 & 34.5 & 39.5 & 41.5  & - & - & - & - & 32.9 & 29.1 & 30.4 & 37.4 & - & - & - & -  \\
T-Rex2 (visual)~\cite{TRex2} & Swin-L & 46.5 & - & 47.6 & 45.4 & 46.0 & 49.5 & - & - & - & - & 45.3 & 43.8 & 42.0 & 49.5 & - & - & - & - \\
T-Rex2 (text)~\cite{TRex2} & Swin-L & 52.2 & - & 54.9 & 49.2 & 54.8 & \underline{56.1} & - & - & - & - & 45.8 & 42.7 & 43.2 & \underline{50.2} & - & - & - & - \\
\midrule
\multicolumn{20}{c}{\textit{Assembled General Perception Model}} \\
\midrule
SAM (ViTDet-H prompt)~\cite{SAM} & - & - & \textcolor{gray}{46.5} & - & - & - & - & - & - & - & - & - & - & - & - & \textcolor{gray}{44.7} & - & - & - \\
Grounded SAM (1.5 Pro + Huge)~\cite{GroundedSAM,SAM} & - &  - & \underline{44.3} & - & - & - & - & \sota{47.7} & \sota{50.2} & \sota{51.7} & \sota{43.8} & - & - & - & - & \sota{41.8} & \sota{46.0} & \sota{42.3} & \sota{39.5} \\
Grounded SAM 2 (1.5 Pro + Large) ~\cite{GroundedSAM,SAM} & - & - & \sota{44.7} & - & - & - & - & \underline{46.2} & \underline{50.1} & \underline{50.1} & \underline{42.0} & - & - & - & - & \underline{40.5} & \underline{44.6} & \underline{41.0} & \underline{38.1} \\
\midrule
\multicolumn{20}{c}{\textit{Object-Centric Vision Model}} \\
\midrule
Grounding DINO 1.5 Pro~\cite{GroundingDINO1.5} & ViT-L & 54.3 & - & 55.7 & 56.1 & 57.5 & 54.1 & - & - & - & - & 47.6 & 44.6 & 47.9 & 48.7 & - & - & - & -  \\
Grounding DINO 1.6 Pro~\cite{GroundingDINO1.5} & ViT-L  & \underline{55.4} & - & \underline{57.7} & \underline{57.5} & \underline{60.5} & 55.3 & - & - & - & - & \underline{51.1} & \underline{51.5} & \sota{52.0} & 50.1 & - & - & - & -  \\
\rowcolor{blue!5} DINO-X Pro & ViT-L   & \sota{56.0} & 37.9 & \sota{59.8} & \sota{63.3} & \sota{61.7} & \sota{57.5} & {43.8} & {46.7} & {47.5} & {40.0} & \sota{52.4} & \sota{56.5} & \underline{51.1} & \sota{51.9} & {38.5} & {44.4} & {38.4} & {36.1} \\
\bottomrule
\end{tabular}
}
\caption{The performance of DINO-X Pro on the COCO, LVIS-minival and LVIS-val benchmarks compared to previous methods. \textcolor{gray}{Gray} numbers indicate that the training dataset includes images or annotations from the COCO or LVIS datasets.}
\label{tab:dino-x-pro-zero-shot}
\end{table} 

\paragraph{Evaluation on Visual-Prompt Based Detection Benchmarks:} To assess the visual prompt object detection capability of DINO-X, we conduct experiments on the few-shot object counting benchmarks. In this task, each test image is accompanied by three visual exemplar boxes representing the target object, and the model is required to output the count of the target object. We evaluate the performance using the FSC147~\cite{ranjan2021learning} and FSCD-LVIS~\cite{nguyen2022few} datasets, which both feature scenes densely populated with small objects. Specifically, FSC147 primarily consists of single-target scenes, where only one type of object is present per image, whereas FSCD-LVIS focuses on multi-target scenes containing multiple object categories. For FSC147, we report the Mean Absolute Error (MAE) metric, and for FSCD-LVIS, we use the Average Precision (AP) metric. Following prior work~\cite{jiang2023t, TRex2}, the visual exemplar boxes are employed as interactive visual prompts. As shown in Table \ref{tab:interactive_visual_prompt}, DINO-X achieves state-of-the-art performance, demonstrating its strong capability in practical visual prompt object detection.

\begin{table}
  \renewcommand{\arraystretch}{1.2}
  \centering
  \scalebox{0.8}{
  \setlength{\tabcolsep}{2.5mm}{
  \begin{tabular}{llccc}
    \toprule
    \multirow{2}{*}{{Type}}& \multirow{2}{*}{{Method}}    & \multicolumn{2}{c} {{FSC147-test}} & {FSCD-LVIS-test}\\ 
    \cmidrule(lr){3-4} &  & MAE & RMSE & AP \\ 
    \midrule
    \multirow{3}{*}{Density Map Regression} &  FamNet~\cite{ranjan2021learning} & 22.1 & 99.5 & \\
    & BMNet+~\cite{shi2022represent} & 14.6 & 91.8 & \\
    & Counting-DETR~\cite{nguyen2022few} & 12.0 & 49.8 & 22.7 \\
    \midrule
    \multirow{3}{*}{Detection} &  T-Rex~\cite{jiang2023t} & \underline{8.72} & - & 40.3 \\
    & T-Rex2~\cite{TRex2} & 10.9 & \underline{36.7} & \underline{43.4} \\
    \rowcolor{blue!5}\cellcolor{white} & DINO-X Pro  & \sota{5.6} & \sota{27.4} & \sota{44.8}\\
    \bottomrule
  \end{tabular}}}
  \vspace{-2mm}
  \caption{The performance of DINO-X Pro on few-shot object counting benchmarks.}
  \label{tab:interactive_visual_prompt}
\end{table}

\subsubsection{Keypoint Detection}

\paragraph{Evaluation on Human 2D Keypoint Benchmarks:} 
We present a comparison of DINO-X with other related works on the COCO~\cite{COCO}, CrowdPose~\cite{shi2022end}, and Human-Art~\cite{ju2023human} benchmarks, as shown in Table~\ref{tab:pose_exp}. We employ the OKS-based Average Precision (AP)~\cite{shi2022end} as the main metrics. Note that the pose head was trained jointly on MSCOCO, CrowdPose, and Human-Art. Hence the evaluation is not a zero-shot setting. But as we froze the backbone of DINO-X and trained only the pose head, the evaluation on object detection and segmentation still follows the zero-shot setting. Training on multiple pose datasets, our model can effectively predicts keypoints across various person styles, including everyday scenarios, crowded environments, occlusions, and artistic representations. While our model achieves an AP that is $1.6$ lower than ED-Pose (primarily due to the limited number of trainable parameters in the pose head), it outperforms existing models on CrowdPose and Human-Art by $3.4$ AP and $1.8$ AP, respectively, showing its remarkable generalization ability on more diverse scenarios. 

\begin{table*}[ht]
\renewcommand{\arraystretch}{1.2}
\centering
\caption{Comparisons with state-of-the-art methods on COCO-val, CrowdPose-test, and Human-Art-val benchmarks. $\dag$ denotes the flipping test. 
The OKS-based Average Precision (AP) is employed as evaluation metric on the datasets.
\textbf{TD}, \textbf{BU}, \textbf{OS}, \textbf{PT} mean
top-down, bottom-up, one-stage and pre-trained methods, respectively. 
}

\resizebox{0.75\linewidth}{!}{
  \begin{tabular}{llccccccccc}
    \toprule
    \multirow{2}*{Method} & \multirow{2}*{Type}
       & \multicolumn{3}{c}{COCO-val} 
       & \multicolumn{3}{c}{CrowdPose-test} 
       & \multicolumn{3}{c}{Human-Art-val}  \\
       \cmidrule(lr){3-5} \cmidrule(lr){6-8} \cmidrule(lr){9-11} & 
       & AP & AP$_{50}$ & AP$_{75}$ 
       & AP & AP$_{50}$ & AP$_{75}$ 
       & AP & AP$_{50}$ & AP$_{75}$  \\
    \midrule
    Sim.Base.\cite{xiao2018simple}$^{\dag}$ & \multirow{2}*{\textbf{TD}} &70.4 &88.6 & 78.3 &60.8&81.4& 65.7&  - &-  & -\\
    HRNet\cite{sun2019deep}$^{\dag}$ &  & 74.4 & 90.5& 81.9& 71.3& 91.1 & 77.5 & 39.9& 54.5 & 42.0 \\ 
    \midrule
    HrHRNet\cite{cheng2020higherhrnet}$^{\dag}$ & \multirow{3}*{\textbf{BU}} & 67.1 & 86.2 &73.0 & 65.9& 86.4 & 70.6 & 34.6 & - & - \\
    DEKR\cite{geng2021bottom}$^{\dag}$  & & 68.0 & 86.7 & 74.5 & 65.7 & 85.7 & 70.4 &  - &-  & -\\
    SWAHR\cite{luo2021rethinking}$^{\dag}$ & & 68.9 & 87.8 & 74.9  & 71.6& 88.5 &77.6 &  - &-  & - \\
        \midrule
    PETR\cite{shi2022end}$^{\dag}$ & \multirow{2}*{\textbf{OS}}  & 64.8 &85.1 &70.2& 71.6& 90.4& 78.3 &  - &-  & -  \\
    
    ED-Pose\cite{yang2023explicit} &   &  \textbf{75.8} & \textbf{92.3} & \textbf{82.9} & \underline{76.6} & \underline{92.4}& \underline{83.3}&  72.3 &-  & -  \\
    \midrule
    \rowcolor{blue!5} DINO-X Pro & \textbf{PT}  &  \underline{74.4} & \underline{90.7}& \underline{81.1}& \textbf{80.0} & \textbf{88.0} & \textbf{84.4} & \textbf{74.1} & \textbf{90.7} & \textbf{81.1}  \\
  \bottomrule
\end{tabular}}
\label{tab:pose_exp}
\end{table*}

\paragraph{Evaluation on Human Hand 2D Keypoint Benchmarks:} In addition to evaluating human pose, we also present hand pose results on the HInt benchmark~\cite{bib:Hamer} with Percentage of Correctly Localized Keypoints (PCK) as the measurement. PCK is a metric used to evaluate the accuracy of keypoint localization. A keypoint is considered correct if the distance between its predicted and ground truth locations is below a specified threshold.  We use a threshold of 0.05 box size, \ie PCK@0.05. During training, we combine the HInt, COCO, and OneHand10K \cite{bib:1h10k} training dataset (a subset of the compared method HaMeR \cite{bib:Hamer}), and evaluate the performance on the HInt test set. As shown in Table~\ref{tab:hint}, DINO-X achieves the best performance on the PCK@0.05 metrics, indicating its strong capability on highly accurate hand pose estimation.

\begin{table*}[ht]
\renewcommand{\arraystretch}{1.2}
\centering
\caption{Comparisons with state-of-the-art methods on HInt dataset. We use PCK@0.05 as the main metrics.
}
\resizebox{0.95\linewidth}{!}{
  \begin{tabular}{ l ccc ccc ccc}
    \toprule
       \multirow{2}*{Method} & \multicolumn{3}{c}{All joints} & \multicolumn{3}{c}{Visible joints} & \multicolumn{3}{c}{Occluded joints} \\
        \cmidrule(lr){2-4} \cmidrule(lr){5-7} \cmidrule(lr){8-10} & New Days & VISOR & Ego4D & New Days & VISOR & Ego4D & New Days & VISOR & Ego4D \\
    \midrule
    FrankMocap \cite{bib:frankmocap} & 16.1 &  16.8 &  13.1 & 20.1 & 20.4 & 16.3 & 9.2 & 11.0 & 8.4 \\
    METRO \cite{bib:Metro} & 14.7 & 16.8 &  13.2 & 19.2 & 19.7 &  15.8 & 7.0 & 10.2 & 8.1 \\
    Mesh Graphormer \cite{bib:MeshGraphormer} & 16.8 & 19.1 &  14.6 & 22.3 & 23.6 &  18.4 &  7.9 & 10.9 &  8.3 \\
    HandOccNet (param) \cite{bib:HandOcc} & 9.1 & 8.1 & 7.7 & 10.2 & 8.5 & 7.3 &7.2 & 7.4 & 8.0 \\
    HandOccNet (no param) \cite{bib:HandOcc} & 13.7 & 12.4 & 10.9 & 15.7 &  13.1 &  11.2 &  9.8 &  9.9 & 9.6 \\
    ViTPose-Hands \cite{bib:vitpose} & 32.2 & 40.0 & 23.3 & 44.0 & 55.7 & 35.0 & 13.9 &  21.2 & 10.3 \\

    Hamba \cite{bib:Hamba} & 48.7 & 47.2 & -- & 61.2 & 61.4 & -- & 28.2 & 29.9 & -- \\
    HaMeR \cite{bib:Hamer} & \underline{51.6} & \underline{56.5} & \underline{46.9} & \underline{62.9} & \underline{66.5} & \underline{59.1} & \underline{33.2} & \underline{42.6} &	\underline{33.1} \\
    \midrule
    \rowcolor{blue!5} DINO-X Pro & \bf 54.3 & \bf 63.0 & \bf 66.0 & \bf 69.3 &	\bf 78.0 &	\bf 81.1 & \bf 34.4 &	\bf 48.0 &	\bf 49.1 \\
  \bottomrule
\end{tabular}}
\label{tab:hint}
\vspace{-0.4cm}
\end{table*}

\subsubsection{Object-Level Vision-Language Understanding}

\paragraph{Evaluation on Object Recognition:} We verify the effectiveness of our language head with related works on object recognition benchmarks, which need to recognize the category of the object in a specified region of an image. Following Osprey\cite{yuan2024osprey}, we use Semantic Similarity (SS) and Semantic IoU (S-IOU)\cite{conti2023vocabulary}, to evaluate the object recognition capability of the language head on the object-level LVIS-val\cite{LVIS} and the part-level PACO-val\cite{ramanathan2023paco} datasets. As shown in Table~\ref{tab:ref_cls}, Our model achieves $71.25\%$ in SS and $41.15\%$ in S-IoU, surpassing Osprey by $6.01\%$ in SS and $2.06\%$ in S-IoU on the LVIS-val dataset. On the PACO dataset, our model is inferior to Osprey. Note that we did not include LVIS and PACO in our langauge head training and the performance of our model is achieved in a zero-shot manner. The lower performance on PACO might be due to the discrepancy between our training data and PACO. And our model only has 1\% trainable parameters compared with Osprey.

\begin{table}
  \renewcommand{\arraystretch}{1.2}
  \centering
  \scalebox{0.8}{
  \setlength{\tabcolsep}{2.5mm}{
  \begin{tabular}{lllcccc}
    \toprule
    \multirow{2}{*}{{Method}}& \multirow{2}{*}{{Visual Encoder}}  & \multirow{2}{*}{{Language Decoder}}  & \multicolumn{2}{c} {{LVIS}} & \multicolumn{2}{c}{{PACO}}\\ 
    \cmidrule(lr){4-5} \cmidrule(lr){6-7}& & & SS & S-IoU & SS & S-IoU \\ 
    \midrule
    Kosmos-2~\cite{peng2023kosmos}  & ViT-L & LM-1.3B~\cite{peng2023kosmos} & 38.95 & 8.67 & 32.09 & 4.79 \\
    Shikra~\cite{chen2023shikra}  & ViT-L & Vicuna-7B\cite{chiang2023vicuna} & 49.65 & 19.82 &  43.64 & 11.42\\
    GPT4RoI~\cite{zhang2023gpt4roi}  & ViT-L & Vicuna-7B\cite{chiang2023vicuna} & 51.32 & 11.99 & 48.04 & 12.08\\
    Ferret~\cite{you2023ferret}  &ViT-L & Vicuna-7B\cite{chiang2023vicuna} & 63.78 & 36.57 & 58.68 & 25.96 \\
    Osprey~\cite{yuan2024osprey}  &ConvNeXt-L & Vicuna-7B\cite{chiang2023vicuna} & \underline{65.24} & \underline{38.19} & \textbf{73.06} & \textbf{52.72} \\
    \midrule
    \rowcolor{blue!5} DINO-X Pro  &ViT-L & OPT-125M\cite{zhang2022opt} & \textbf{71.25} & \textbf{41.15} & \underline{66.67} & \underline{39.39}\\
    \bottomrule
  \end{tabular}}}
  \vspace{-2mm}
  \caption{Results on referring object classification benchmarks. We use Semantic Similarity (SS) and Semantic-IoU (S-IoU) scores to measure the region classification quality.}
  \label{tab:ref_cls}
\end{table}

\paragraph{Evaluation on Region Captioning:} We evaluate our model's region caption quality on Visual Genome\cite{krishna2017visual} and RefCOCOg\cite{mao2016generation}. The evaluation results are presented in Table~\ref{tab:reion_cap}. Remarkably, based on object-level features extracted by a frozen DINO-X backbone and without utilizing any Visual Genome training data, our model achieves a $142.1$ CIDEr score on the Visual Genome benchmark in a zero-shot manner. Further, after fine-tuning on Visual Genome dataset, we set a new state-of-the-art result with $201.8$ CIDEr score with only a light-weight language head.

\begin{table}
  \renewcommand{\arraystretch}{1.2}
  \centering
  \scalebox{0.75}{
  \setlength{\tabcolsep}{2.5mm}{
  \begin{tabular}{lllcccc}
    \toprule
    \multirow{2}{*}{{Method}} & \multirow{2}{*}{{Visual Encoder}}  & \multirow{2}{*}{{Langauge Decoder}}  & \multicolumn{2}{c} {{Visual Genome}} & \multicolumn{2}{c}{{RefCOCOg}}\\ 
    \cmidrule(lr){4-5} \cmidrule(lr){6-7}& & &\makecell{CIDEr} & \makecell{METEOR} &\makecell{CIDEr} & \makecell{METEOR} \\ 
    \midrule
    GRIT~\cite{wu2025grit} & ViT-B & Small-43M~\cite{wu2025grit}  & 142.0 & 17.2 & 71.6 & 15.2 \\
    GPT4RoI~\cite{zhang2023gpt4roi} & ViT-L & Vicuna-7B\cite{chiang2023vicuna} & 145.2 & 17.4 & - & -\\
    ASM~\cite{wang2023all} & ViT-G & Husky-7B\cite{kim2024husky} & 145.1 & 18.0 & \underline{103.0} & \sota{20.8}\\ 
    AlphaCLIP~\cite{sun2024alpha} & ViT-L & Vicuna-7B\cite{chiang2023vicuna} & \underline{160.3} & \underline{18.9} & \sota{109.2} & \underline{16.7} \\
    SCA~\cite{huang2024segment} & SAM-H & Llama-3B\cite{dubey2024llama} & 149.8 & 17.4 & 74.0 & 15.6 \\
    \midrule
    \rowcolor{blue!5} DINO-X Pro \tiny{(zero-shot)} & ViT-L & OPT-125M\cite{zhang2022opt} & 143.2 & 17.5 & 55.7 & 12.2\\
    \rowcolor{blue!5} DINO-X Pro \tiny{(fine-tuned)} & ViT-L & OPT-125M\cite{zhang2022opt} & \textbf{201.8} & \textbf{20.1} & 86.3 & 15.1\\
    \bottomrule
  \end{tabular}}}
  \vspace{-2mm}
  \caption{Results on region captioning benchmarks. We report METEOR and CIDEr scores to measure the region caption quality.}
  \label{tab:reion_cap}
\end{table}

\subsection{DINO-X Edge}
\label{dinox_edge_eval_results}

\begin{table}[!h]
\setlength{\tabcolsep}{3pt}
\renewcommand{\arraystretch}{1.25}
\centering
\scriptsize
\resizebox{1.0\linewidth}{!}{
\begin{tabular}{lccccccccccccc}
\toprule
\multirow{2}*{Method} & \multirow{2}*{Backbone}  & \multirow{2}*{Test Size} & COCO-val & \multicolumn{4}{c}{LVIS-minival} & \multicolumn{4}{c}{LVIS-val} & FPS (A100) & FPS (Orin NX)\\
\cmidrule(lr){5-8} \cmidrule(lr){9-12} & & & $\text{AP}_{box}$ & $\text{AP}_{all}$ & $\text{AP}_{r}$ & $\text{AP}_{c}$ & $\text{AP}_{f}$ & $\text{AP}_{all}$ & $\text{AP}_{r}$ & $\text{AP}_{c}$ & $\text{AP}_{f}$ & Pytorch/TensorRT FP32 & TensorRT FP32/FP16\\
\midrule
\multicolumn{14}{c}{\textit{End-to-End Open-Set Object Detection}}\\
\midrule
GLIP~\cite{GLIP} & Swin-T~\cite{SwinTransformer} & 800 $\times$ 1333 & 46.3 & 26.0 & 20.8 & 21.4 & 31.0 & - & - & - & - & - & -/- \\
Grounding DINO~\cite{GroundingDINO} & Swin-T~\cite{SwinTransformer}  & 800 $\times$ 1333 & 48.4 & 27.4 & 18.1 & 23.3 & 32.7 & - & - & - & - &  9.4 / 42.6 & 1.1/- \\ 
\midrule
\multicolumn{14}{c}{\textit{Real-time End-to-End Open-Set Object Detection Models}} \\
\midrule
YOLO-Worldv2-S\dag~\cite{YOLO-World}  & YOLOv8-S~\cite{YOLOv8}  & 640 $\times$ 640 & - & 22.7 & 16.3 & 20.8 & 25.5 & 17.3 & 11.3 & 14.9	& 22.7 & 47.4 / - & -/- \\
YOLO-Worldv2-M\dag~\cite{YOLO-World}  & YOLOv8-M~\cite{YOLOv8}  & 640 $\times$ 640 & - & 30.0 & 25.0 & 27.2 & 33.4 & 23.5 & 17.1 & 20.0	& 30.1 & 42.7 / - & -/- \\
YOLO-Worldv2-L\dag~\cite{YOLO-World}  & YOLOv8-L~\cite{YOLOv8} & 640 $\times$ 640 & - & 33.0	& 22.6 & 32.0 & 35.8 & 26.0 & 18.6 & 23.0 & 32.6 & 37.4 / - & -/- \\
YOLO-Worldv2-L\dag~\cite{YOLO-World}  & YOLOv8-L~\cite{YOLOv8} & 640 $\times$ 640 & - & 32.9	& 25.3 & 31.1 & 35.8 & 26.1 & 20.6 & 22.6 & 32.3 & 37.4 / - & -/- \\
OmDet-Turbo-T~\cite{OmDet_Turbo}  & Swin-T~\cite{SwinTransformer}  & 640 $\times$ 640 & 42.5 & 30.3 & - & - & - & - & - & - & - & 21.5 / 140.0 & -/- \\
OVLW-DETR-L~\cite{wang2024ovlw}  & LW-DETR-L~\cite{chen2024lw} & 640 $\times$ 640 & - & 33.5	& 26.5 & 33.9 & 34.4 & - & - & - & - & - / - & -/- \\
\midrule
\multicolumn{14}{c}{\textit{Efficient Object-Centric Vision Model}}\\
\midrule
Grounding DINO 1.5 Edge~\cite{GroundingDINO1.5}& EfficientViT-L1~\cite{cai2023efficientvit}  & 640 $\times$ 640 & 42.9 & 33.5 & 28.0 & 34.3 & 33.9 & 27.3 & 26.3 & 25.7 & 29.6 & 21.7 / 111.6 & 10.7/- \\ 
Grounding DINO 1.5 Edge~\cite{GroundingDINO1.5}& EfficientViT-L1~\cite{cai2023efficientvit}  &  800 $\times$ 1333 & 45.0 & 36.2 & 33.2 & 36.6 & 36.3 & 29.3 & 28.1 & 27.6 & 31.6 & 18.5 / 75.2 & 5.5/- \\ 
Grounding DINO 1.6 Edge~\cite{GroundingDINO1.5} & EfficientViT-L1~\cite{cai2023efficientvit} &  800 $\times$ 800 & 44.8 & 36.9 & 34.6 & 39.1 & 35.4 & 31.0 & 31.6 & 30.5  & 31.4  &  20.81/152.7  & 10.0/15.1 \\ 
Grounding DINO 1.6 Edge~\cite{GroundingDINO1.5} & EfficientViT-L1~\cite{cai2023efficientvit}  &  1024 $\times$ 1024 & 46.5 & 40.1 & 36.8 & 42.0 & 39.0 & 33.3 & 32.6 & 32.8  & 34.3  &  19.4/108.1  & 7.6/10.5 \\ 
\rowcolor{blue!5} {DINO-X Edge} & EfficientViT-L2~\cite{cai2023efficientvit}  &  640 $\times$ 640 & 48.7 & \underline{44.5} & \underline{41.4} & \underline{47.3} & \underline{42.6} & \underline{38.4} & \underline{38.9} & \underline{38.3}  & \underline{38.2}  &  19.8/138.6  & 10.0/20.1 \\ 
\rowcolor{blue!5} {DINO-X Edge} & EfficientViT-L2~\cite{cai2023efficientvit}  &  800 $\times$ 1333 & \textbf{50.9} & \textbf{48.3} & \textbf{47.6} & \textbf{50.2} & \textbf{46.6} & \textbf{42.0} & \textbf{43.1} & \textbf{41.7}  & \textbf{41.8}  &  15.1/74.5  & 4.5/9.1 \\ 
\bottomrule
\end{tabular}}
\caption{Zero-shot Performance of DINO-X Edge on COCO, LVIS-minival, and LVIS-val object detection benchmarks compared with related works.}
\label{tab:edge}
\end{table} 

\paragraph{Evaluation on Zero-Shot Object Detection Benchmarks:}
To evaluate the zero-shot object detection capability of DINO-X Edge, we conduct tests 
on the COCO and LVIS benchmarks after pre-training on Grounding-100M.
As shown in Table~\ref{tab:edge}, DINO-X Edge outperforms existing real-time open-set detectors on COCO benchmark by a large margin. DINO-X Edge also achieves $48.3$ AP and $42.0$ AP on LVIS-minival and LVIS-val, respectively, demonstrating excellent zero-shot detection capability in long-tailed detection scenarios.

We evaluate the inference speed DINO-X Edge using both FP32 and FP16 TensorRT models on NVIDIA Orin NX, measuring the performance in terms of frames per second (FPS). The FPS results for the PyTorch model and the FP32 TensorRT model on an A100 GPU were also included. \dag denotes that the YOLO-World results were reproduced using the latest official codes.

Leveraging the normalization technique in floating-point multiplication, we can quantize the model to FP16 without sacrificing the performance. With an input size of 640×640, DINO-X Edge achieves an inference speed of $20.1$ FPS, marking a $33\%$ improvement compared to Grounding DINO 1.6 Edge (increasing from $15.1$ FPS to $20.1$ FPS).

\section{Case Analysis and Qualitative Visualization}

In this section, we visualize the different capabilities of DINO-X models across various real-world scenarios. The images are primarily sourced from COCO~\cite{COCO}, LVIS~\cite{LVIS}, V3Det~\cite{V3Det}, SA-1B~\cite{SAM}, and other publicly available resources. We are deeply grateful for their contributions, which have significantly benefited the community.

\subsection{Open-World Object Detection}

\begin{figure}[ht!]
\centering
 \includegraphics[width=1.0\textwidth,keepaspectratio]{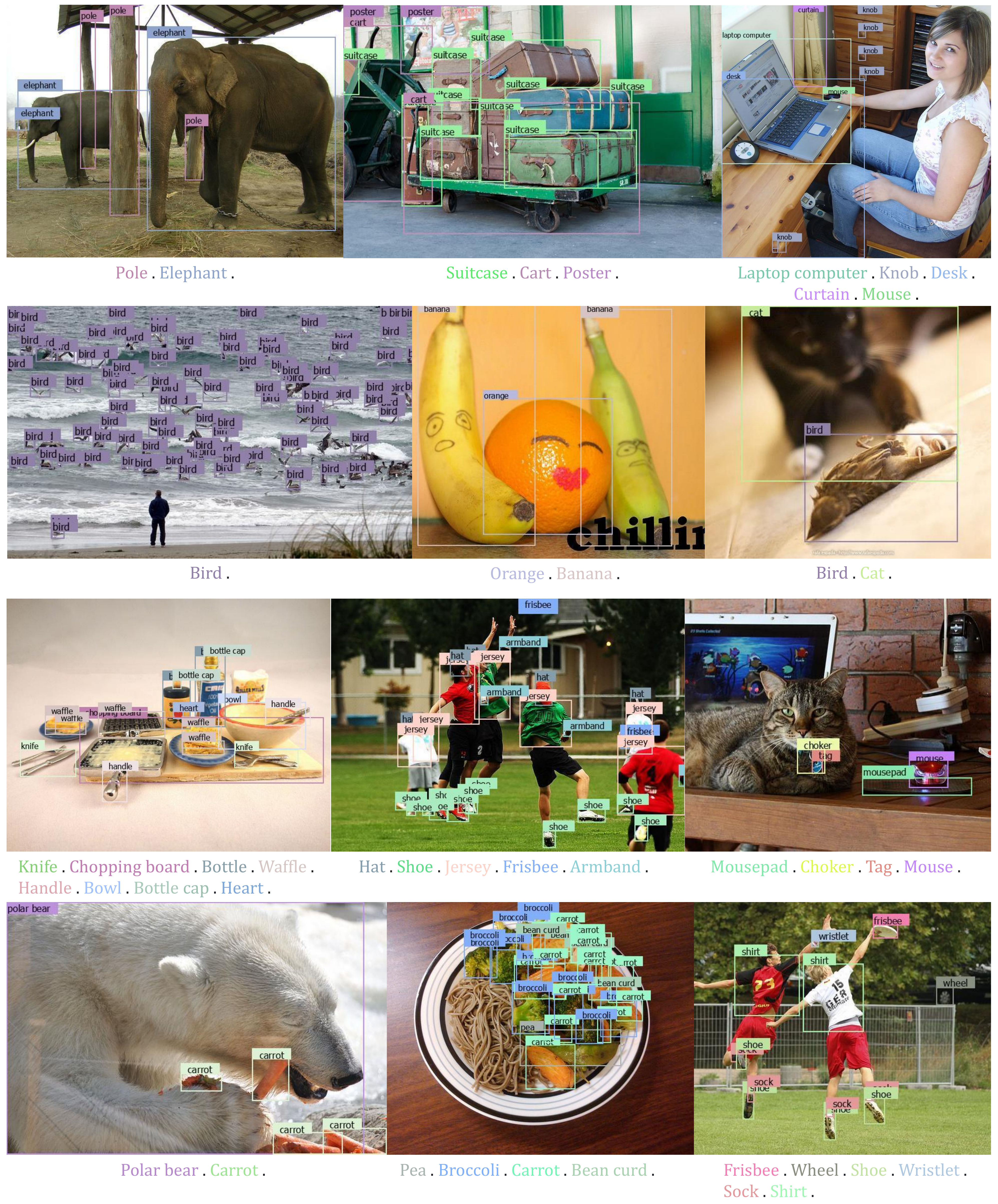}
 \caption{Open-world object detection with DINO-X}
 \label{fig:open_world_detection}
\end{figure}

As illustrated in Figure~\ref{fig:open_world_detection}, DINO-X demonstrates the capability to detect any objects based on the given text prompt. It can identify a wide range of objects, from common categories to long-tailed classes and dense object scenarios, showcasing its robust open-world object detection capabilities.

\subsection{Long Caption Phrase Grounding}

\begin{figure}[ht!]
\centering
 \includegraphics[width=1.0\textwidth,keepaspectratio]{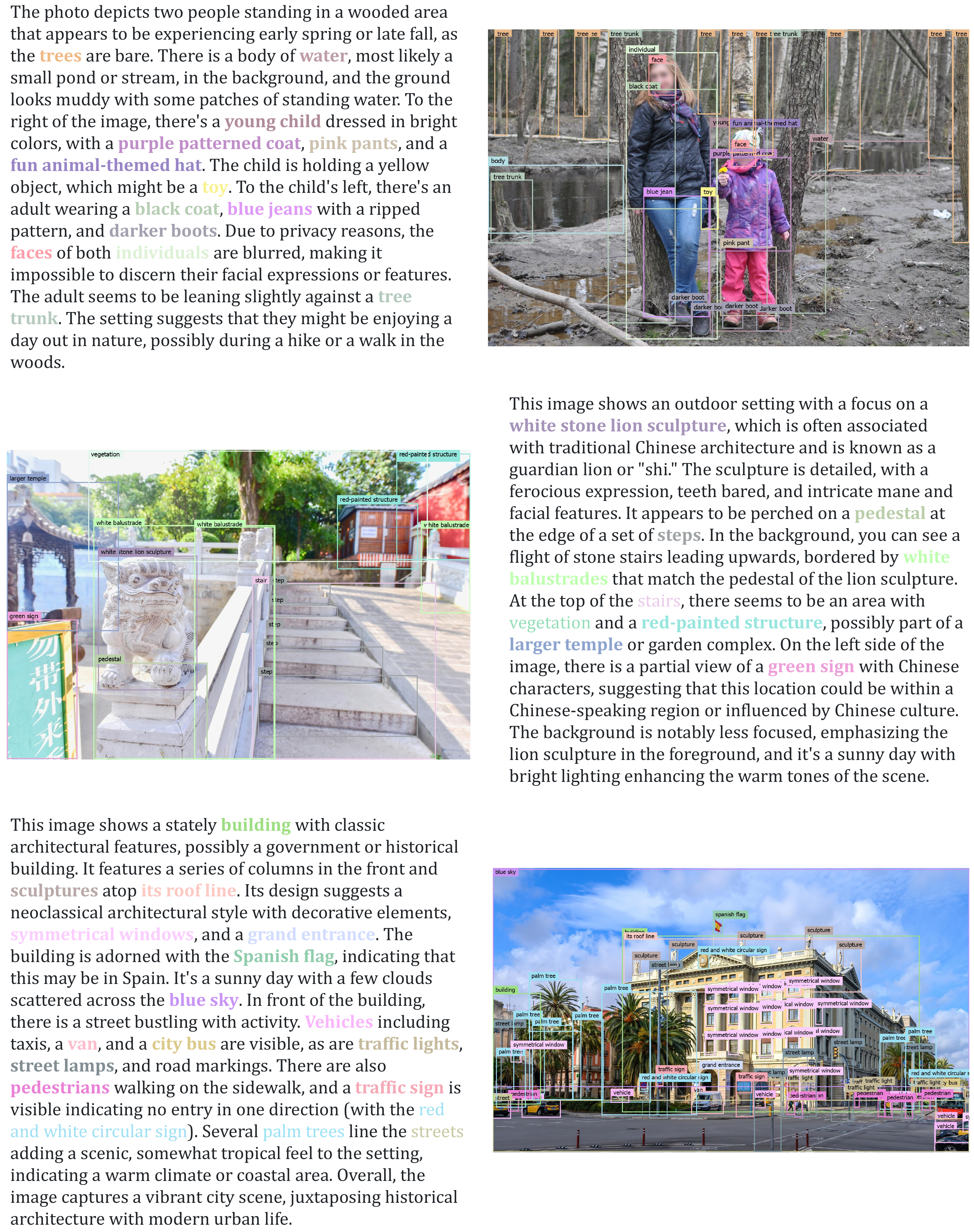}
 \caption{Long caption phrase grounding with DINO-X}
 \label{fig:phrase_grounding}
\end{figure}

As illustrated in Figure~\ref{fig:phrase_grounding}, DINO-X exhibits an impressive ability to locate corresponding regions in an image based on noun phrases from a long caption. The capability of mapping each noun phrase in a detailed caption to specific objects in an image marks a significant advancement in deep image understanding. This feature has substantial practical value, such as enabling multimodal large language models (MLLMs) to generate more accurate and reliable responses.

\clearpage

\subsection{Open-World Object Segmentation and Visual Prompt Counting}

\begin{figure}[ht!]
\centering
 \includegraphics[width=1.0\textwidth,keepaspectratio]{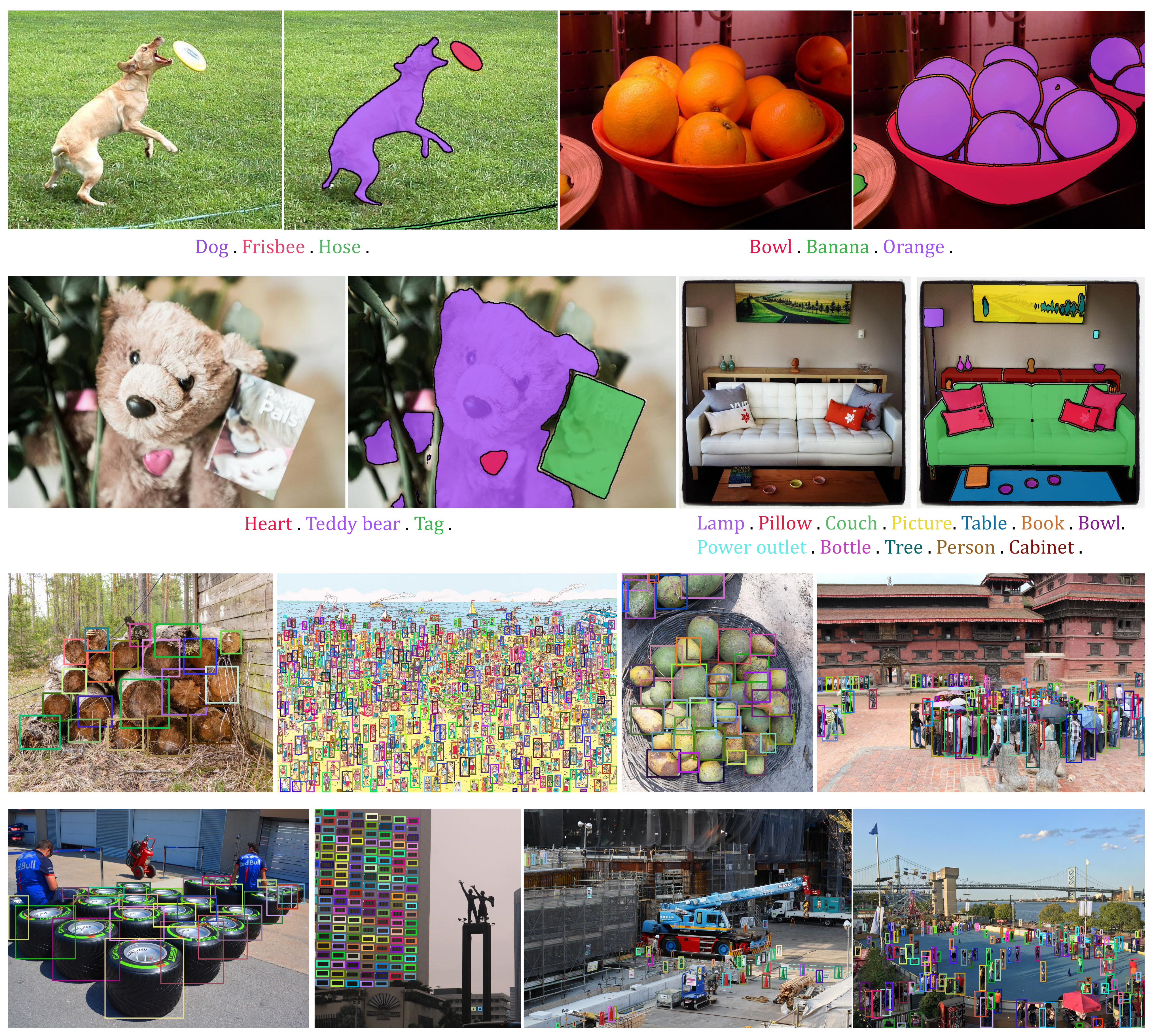}
 \caption{Open-world object segmentation and visual prompt object counting with DINO-X}
 \label{fig:open_world_segmentation_visual_prompt_counting}
\end{figure}

As shown in Figure~\ref{fig:open_world_segmentation_visual_prompt_counting}, beyond Grounding DINO 1.5~\cite{GroundingDINO1.5}, DINO-X not only enables open-world object detection based on text prompts but also generates the corresponding segmentation mask for each object, providing richer semantic outputs. Furthermore, DINO-X also supports detection based on user-defined visual prompts by drawing bounding boxes or points on target objects. This capability demonstrates exceptional usability in object counting scenarios.

\clearpage

\subsection{Prompt-Free Object Detection and Recognition}
\begin{figure}[ht!]
\centering
 \includegraphics[width=1.0\textwidth,keepaspectratio]{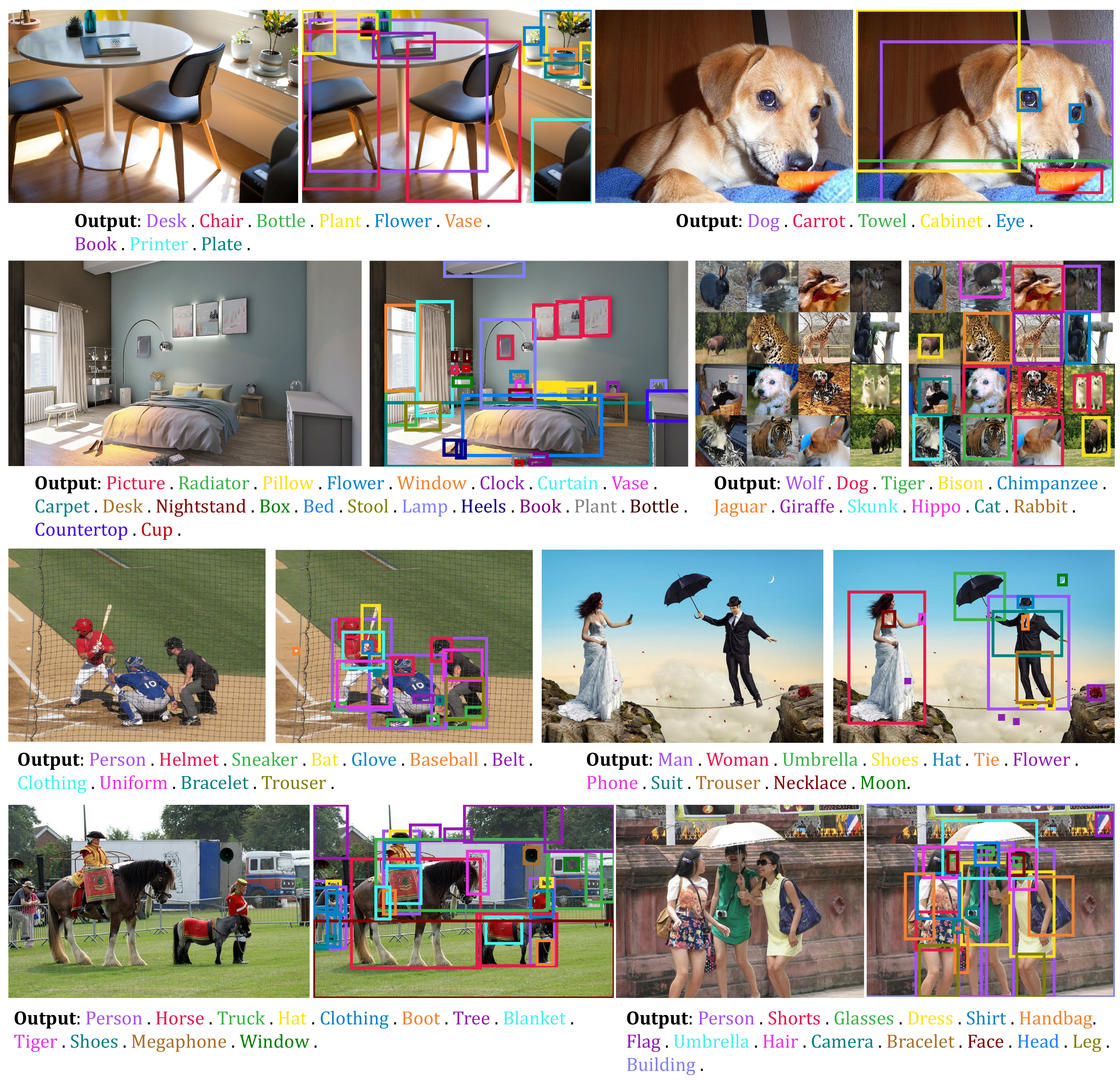}
 \caption{Prompt-free object detection and recognition with DINO-X}
 \label{fig:prompt_free_detection_and_recognition}
\end{figure}

In DINO-X, we developed a highly practical feature named \textit{prompt-free} object detection, which allows users to detect any objects in an input image without providing any prompts. As shown in Figure~\ref{fig:prompt_free_detection_and_recognition} When combined with DINO-X's language head, this feature enables seamless detection and identification of all objects in the image without requiring any user input.

\clearpage

\subsection{Dense Region Caption}

\begin{figure}[ht!]
\centering
 \includegraphics[width=1.0\textwidth,keepaspectratio]{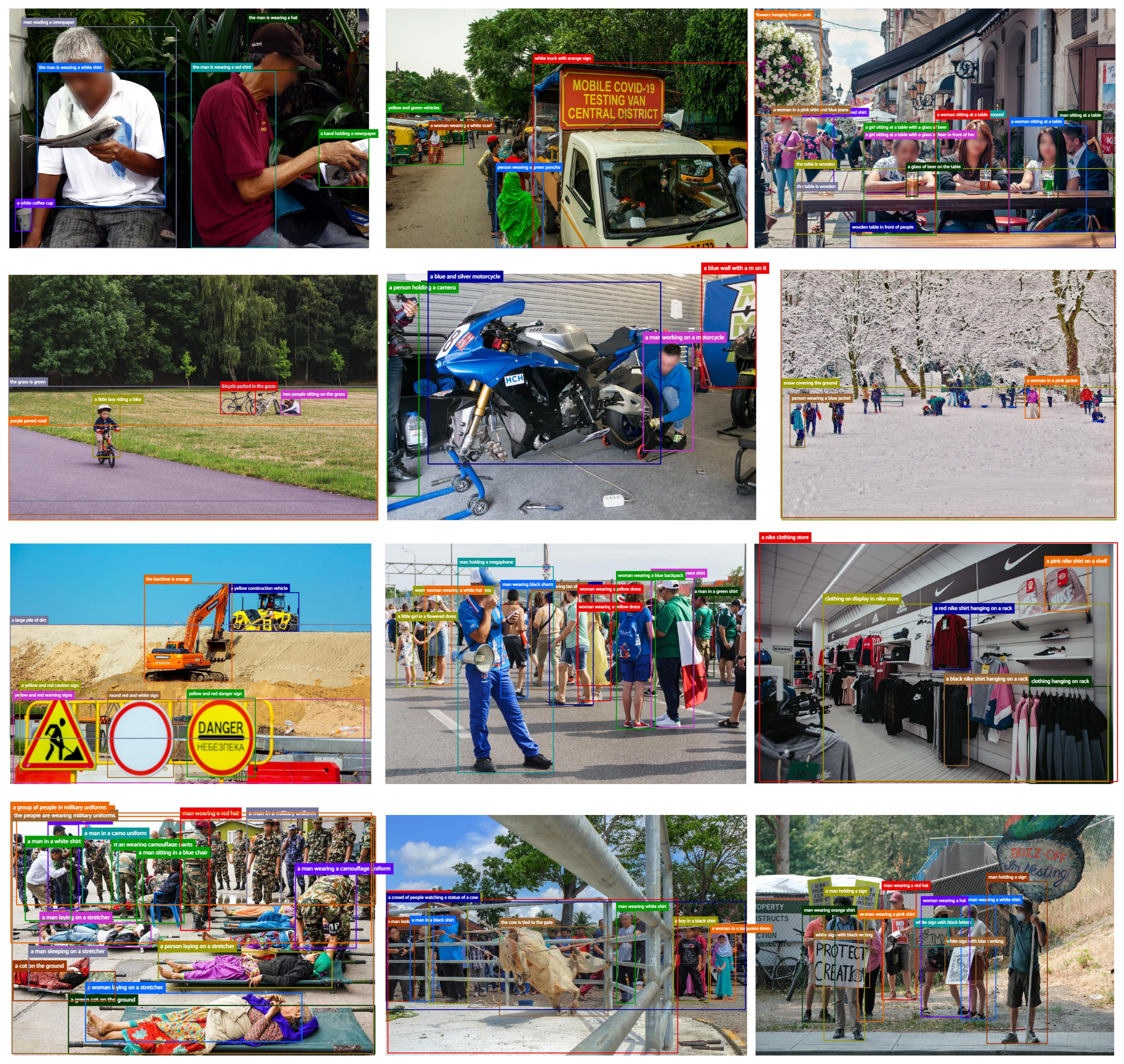}
 \caption{Dense Region Caption with DINO-X}
 \label{fig:dense_region_caption}
\end{figure}

As illustrated in Figure~\ref{fig:dense_region_caption}, DINO-X can generate more fine-grained captions for any specified region. Furthermore, with DINO-X's language head, we can also perform tasks such as region-based QA and other region understanding tasks. Currently, this feature is still in the development stage and will be released in our next version.

\clearpage

\subsection{Human Body and Hand Pose Estimation}

\begin{figure}[ht!]
\centering
 \includegraphics[width=1.0\textwidth,keepaspectratio]{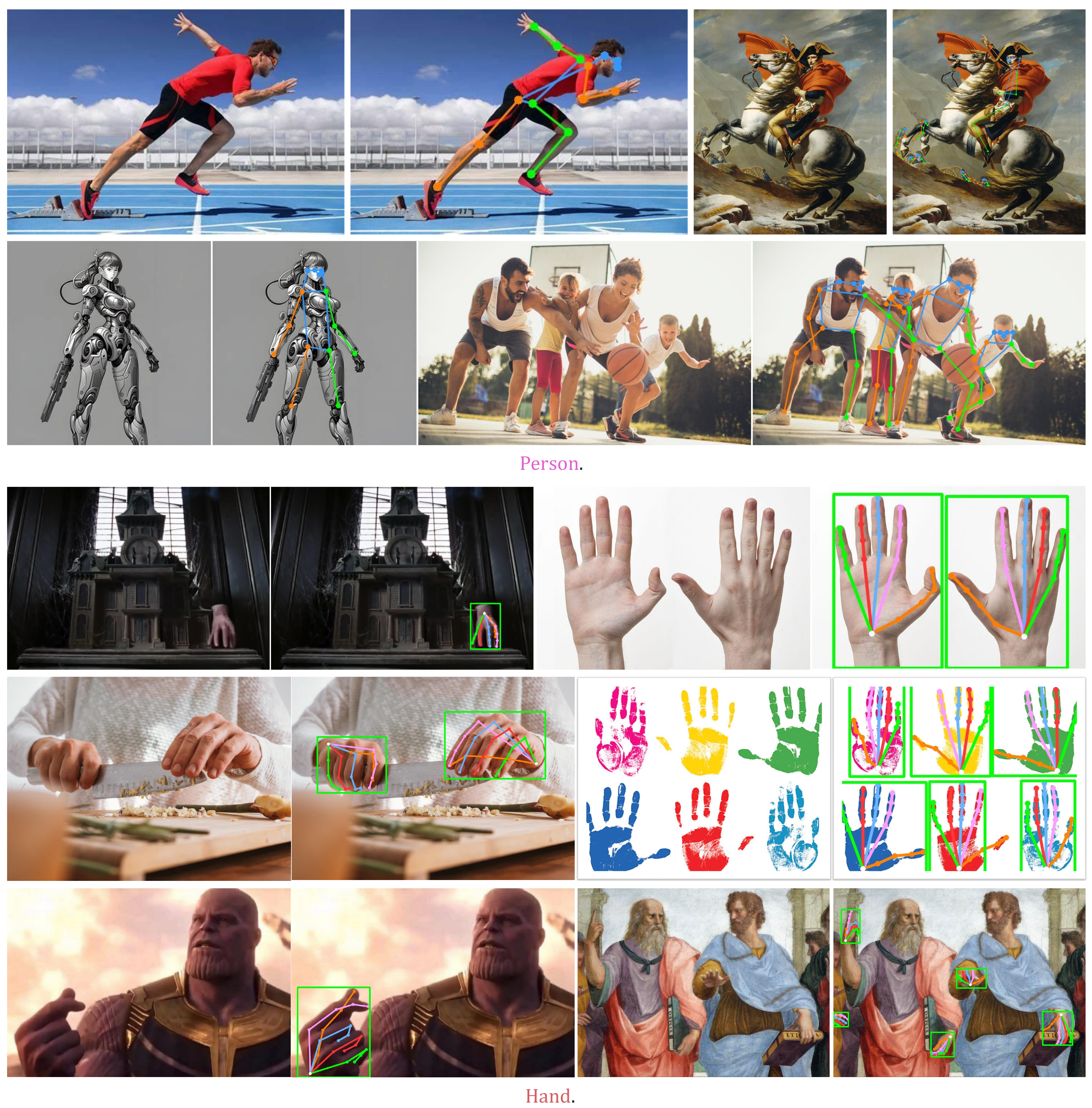}
 \caption{Pose estimation on human body and human hand with DINO-X}
 \label{fig:human_body_hand_vis}
\end{figure}

As shown in Figure~\ref{fig:human_body_hand_vis}, DINO-X can predict keypoints for specific categories through the keypoint heads based on the text prompts. Trained on a combination of COCO, CrowdHuman, and Human-Art datasets, DINO-X is capable of predicting human body and hand keypoints across various scenarios.

\clearpage

\subsection{Side-by-side comparison with Grounding DINO 1.5 Pro}

\begin{figure}[ht!]
\centering
 \includegraphics[width=1.0\textwidth,keepaspectratio]{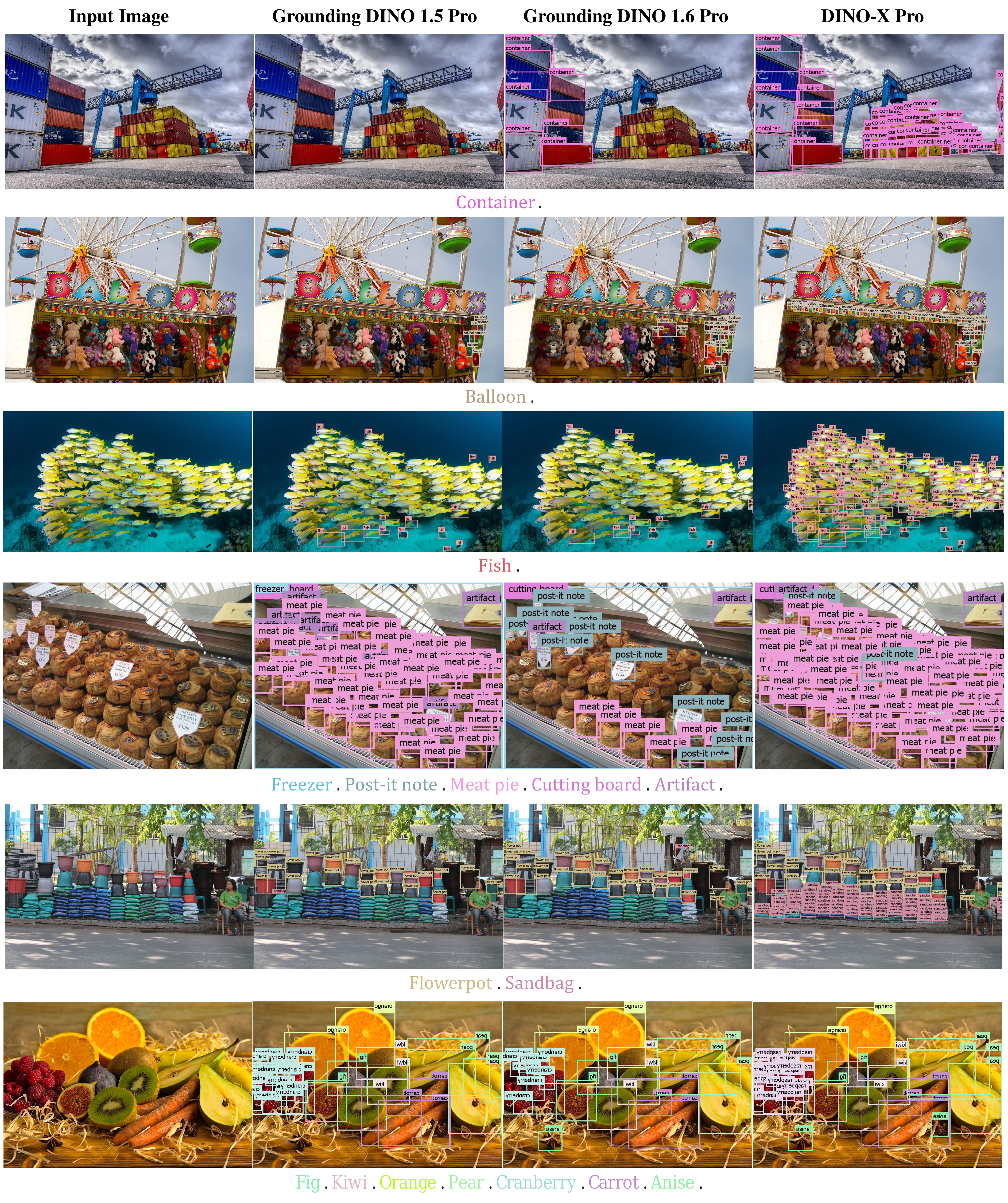}
 \caption{Comparison of Grounding DINO 1.5 Pro, Grounding DINO 1.6 Pro and DINO-X}
 \label{fig:dinox_compare_gd1.5}
\end{figure}

We conducted a side-by-side comparison of DINO-X with previous state-of-the-art models, Grounding DINO 1.5 Pro and Grounding DINO 1.6 Pro. As shown in Figure~\ref{fig:dinox_compare_gd1.5}, built upon the foundation of Grounding DINO 1.5, DINO-X further enhances its language comprehension capabilities while delivering a remarkable performance in dense object detection scenarios.

\clearpage
\section{Conclusion}
This paper has presented DINO-X, a strong object-centric vision model to advance the field of open-set object detection and understanding. 
The flagship model, DINO-X Pro, has established new records on the COCO and LVIS zero-shot benchmarks, showing a remarkable improvement in detection accuracy and reliability. To make long-tailed object detection easy, DINO-X not only supports open-world detection based on text prompts but also enables object detection with visual prompts and customized prompts for customized scenarios. Moreover, DINO-X extends its capabilities from detection to a broader range of perception tasks, including segmentation, pose estimation, and object-level understanding tasks. To enable real-time object detection for more applications on edge devices, we also developed the DINO-X Edge model, which further expands the practical utility of the DINO-X series models.
\section{Contributions and Acknowledgments}
We would like to express our gratitude to everyone involved in the DINO-X project. The contributions are as follows (in no particular order):

\vspace{-5pt}
\begin{itemize}[leftmargin=*]
    \item \textbf{DINO-X Pro}: Yihao Chen, Tianhe Ren, Qing Jiang, Zhaoyang Zeng, and Yuda Xiong.
    \item \textbf{Mask Head}: Tianhe Ren, Hao Zhang, Feng Li, and Zhaoyang Zeng.
    \item \textbf{Visual Prompt \& Prompt-Free Detection}: Qing Jiang.
    \item \textbf{Pose Head}: Xiaoke Jiang, Xingyu Chen, Zhuheng Song, and Yuhong Zhang.
    \item \textbf{Language Head}: Wenlong Liu, Zhengyu Ma, Junyi Shen, Yuan Gao, and Yuda Xiong.
    \item \textbf{DINO-X Edge}: Hongjie Huang, Han Gao, and Qing Jiang.
    \item  \textbf{Grounding-100M}: Yuda Xiong, Yihao Chen, Tianhe Ren, Qing Jiang, Zhaoyang Zeng, and Shilong Liu.
    \item \textbf{Language Head and DINO-X Edge Lead}: Kent Yu.
    \item \textbf{Overall Project Lead}: Lei Zhang.
\end{itemize}

We would also like to thank everyone involved in the DINO-X playground and API support, including application lead Wei Liu, product manager Qin Liu and Xiaohui Wang, front-end developers Yuanhao Zhu, Ce Feng, and Jiongrong Fan, back-end developers Zhiqiang Li and Jiawei Shi, UX designer Zijun Deng, operation intern Weijian Zeng, tester Jiangyan Wang, and Peng Xiao for providing suggestions and feedbacks on customized scenarios.


\clearpage
\bibliographystyle{plain}
\bibliography{main,caption}





\end{document}